\documentclass[letterpaper, 10 pt, conference]{ieeeconf}  
\IEEEoverridecommandlockouts
\usepackage{cite}
\usepackage{amsmath,amssymb,amsfonts}
\usepackage{algorithm}
\usepackage[noend]{algpseudocode}
\algrenewcommand\algorithmicindent{0.7em}%

\DeclareMathOperator{\svd}{SVD}
\newcommand\Algphase[1]{%
\vspace*{-.7\baselineskip}\Statex\hspace*{\dimexpr-\algorithmicindent-2pt\relax}\rule{0.97\columnwidth}{0.2pt}%
\Statex\hspace*{-\algorithmicindent}\textbf{#1}%
\vspace*{-.7\baselineskip}\Statex\hspace*{\dimexpr-\algorithmicindent-2pt\relax}\rule{0.97\columnwidth}{0.2pt}%
}
\newcommand\True{\textbf{true}}
\newcommand\False{\textbf{false}}
\usepackage{graphicx}
\usepackage{textcomp}
\usepackage{xcolor}
\usepackage{svg}
\usepackage[nolist]{acronym}
\usepackage[font=small]{caption}
\usepackage{units}
\usepackage{comment}
\usepackage{booktabs}
\usepackage{tikz}
\usepackage{hyperref}
\usepackage{subfig}
\usepackage{microtype}
\usepackage{geometry}
\usetikzlibrary{shapes.geometric, arrows, calc,patterns,angles,quotes}
\usetikzlibrary{calc}
\usetikzlibrary{positioning}
\tikzstyle{process} = [rectangle, minimum width=2.5cm, minimum height=1cm, text centered, draw=black]
\tikzstyle{arrow} = [thick,->,>=stealth]
\tikzstyle{process} = [rectangle, minimum width=3.0cm, minimum height=3.0cm, text centered, text width=3.0cm, draw=black]
\tikzstyle{process_content} = [rectangle, minimum width=3.0cm, text centered, text width=3.0cm, yshift=-0.3cm]

\newcommand{\etal}{\textit{et al}. }

\usepackage[normalem]{ulem} 

\newcommand{\reviewchanges}[1]{{#1}} 
\newcommand{\finalsubmission}[1]{{#1}} 
\renewcommand{\sout}[1]{}

\begin{document}

\title{\finalsubmission{\sout{HoughCeption}Hough$^2$Map} -- Iterative Event-based Hough Transform for High-Speed Railway Mapping}
\author{Florian Tschopp$^{1,\ast}$, Cornelius von Einem$^{1,2,\ast}$, Andrei Cramariuc$^{1,\ast}$, David Hug$^2$, Andrew W. Palmer$^3$,\\ Roland Siegwart$^1$, Margarita Chli$^2$, and Juan Nieto\finalsubmission{\sout{$^1$}$^{4}$}%
\thanks{$^\ast$Authors contributed equally to this work}%
\thanks{$^1$Authors are members of the Autonomous Systems Lab, ETH Zurich, Switzerland; {\tt\small \{firstname.lastname\}@mavt.ethz.ch}}%
\thanks{$^2$Authors are members of the Vision for Robotics Lab, ETH Zurich, Switzerland; {\tt\small \{firstname.lastname\}@mavt.ethz.ch}}%
\thanks{$^3$Author \reviewchanges{\sout{is}was} with Siemens Mobility, Berlin, Germany; {\tt\small andrew.palmer@\reviewchanges{\sout{siemens.com}emesent.io}}}%
\thanks{\finalsubmission{$^4$Author is with Microsoft, Switzerland but the work was done while the author was a member of $^1$; {\tt\small juannieto@microsoft.com}}}%
\thanks{This work was supported by Siemens Mobility GmbH, Germany and the ETH Mobility Initiative under the project \textit{PROMPT}. The code \finalsubmission{\sout{will be made available upon acceptance}is available at} \url{https://github.com/ethz-asl/Hough2Map}} %
\thanks{\copyright 2021 IEEE.  Personal use of this material is permitted.  Permission from IEEE must be obtained for all other uses, in any current or future media, including reprinting/republishing this material for advertising or promotional purposes, creating new collective works, for resale or redistribution to servers or lists, or reuse of any copyrighted component of this work in other works.}
}

\maketitle
\thispagestyle{empty}
\pagestyle{empty}

\begin{abstract}
To cope with the growing demand for transportation on the railway system, accurate, robust, and high-frequency positioning is required to enable a safe and efficient utilization of the existing railway infrastructure. 
As a basis for a\reviewchanges{\sout{ future}} localization system we propose a complete on-board mapping pipeline able to \reviewchanges{\sout{collect}map} robust meaningful landmarks\reviewchanges{, such as poles from power lines,} in the \reviewchanges{vicinity of the vehicle\sout{, such as poles from power lines or signals}}.
Such poles are good candidates for reliable and long term landmarks even through difficult weather conditions or seasonal changes. 
To address the challenges of motion blur and illumination changes in railway scenarios we employ a \acl{DVS}, a novel event-based camera.
Using a sideways oriented on-board camera, poles appear as vertical lines.
To map such lines in a real-time event stream, we introduce \finalsubmission{\sout{HoughCeption}Hough$^\mathbf{2}$Map}, a novel consecutive iterative event-based Hough transform framework capable of detecting, tracking, and triangulating close-by structures.
We demonstrate the mapping reliability and accuracy of \finalsubmission{\sout{HoughCeption}Hough$^\mathbf{2}$Map} on real-world data in typical usage scenarios and evaluate using surveyed infrastructure ground truth maps.
\finalsubmission{\sout{HoughCeption}Hough$^\mathbf{2}$Map} achieves a detection reliability of up to  $\unit[92]{\%}$ and a mapping root mean square error accuracy of $\unit[1.1518]{m}$.
\end{abstract}

\section{Introduction}
A steady increase in global trade, mobility needs as well as in the environmental consciousness of consumers lead to a heavy demand on rail transportation.
Current railway infrastructures are reaching their operational limits necessitating future-proof utilization schemes with increased efficiency.
Most presently deployed railway safety systems, such as the \ac{ETCS} Level 0-2~\cite{Stanley2011}, operate on fixed block interlocking relying on track-side infrastructure beacons (Balises) which inherently limits the number of deployable trains in addition to high maintenance costs.
%
%
However, transitioning into a more efficient moving block strategy requires accurate and continuous knowledge of \reviewchanges{the position of a train}~\cite{Beugin2012Simulation-basedLocalization,Marais2017ASignaling,Albrecht2013}, which, ideally, should be purely based on on-board sensors to avoid the need for external infrastructure~\cite{Albrecht2013}.

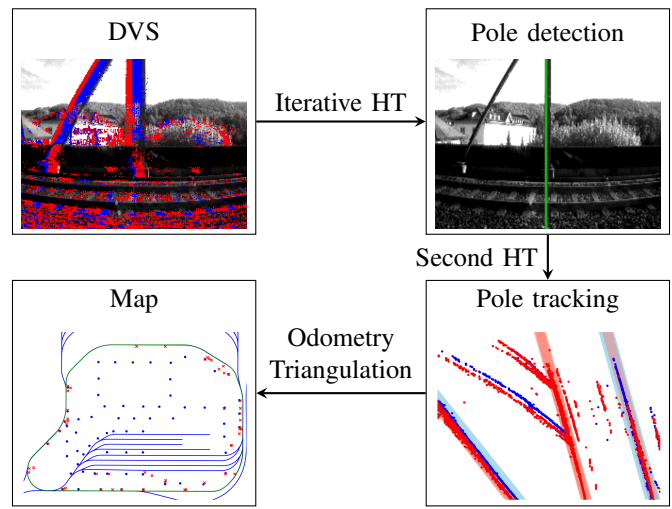
\begin{figure}[ht!]
    \centering
    \begin{tikzpicture}[node distance=5.5cm]
        \coordinate (A) at (1.5,0);
    
        \node (start) [process] {};
        \node (pro1) [process, right of=start] {};
        \node (pro2) [process, below=1.1 of pro1,yshift=0.5cm] {};
        \node (stop) [process, left of=pro2] {};

        \node (dvs_image) [process_content,label=above:DVS]  at (start.center) {
        \includegraphics[width=\textwidth, trim={0cm 0cm 0cm 0cm}, clip]{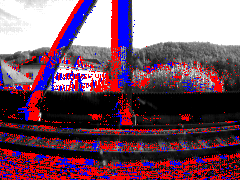}};
        \node (detection_image) [process_content,label=above:Pole detection]  at (pro1.center) {
        \includegraphics[width=\textwidth,trim={0cm 0cm 0cm 0cm},clip]{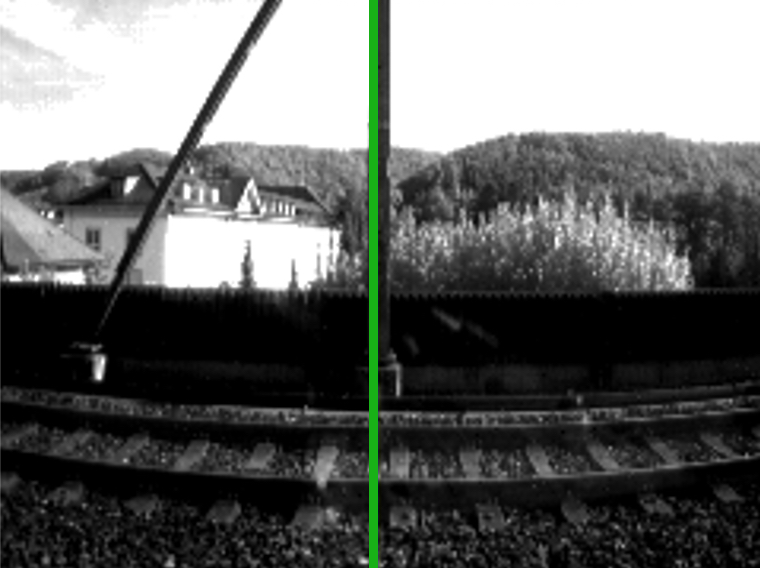}};
        \node (tracking_image) [process_content,label=above:Pole tracking]  at (pro2.center) {
        \includegraphics[width=\textwidth, clip]{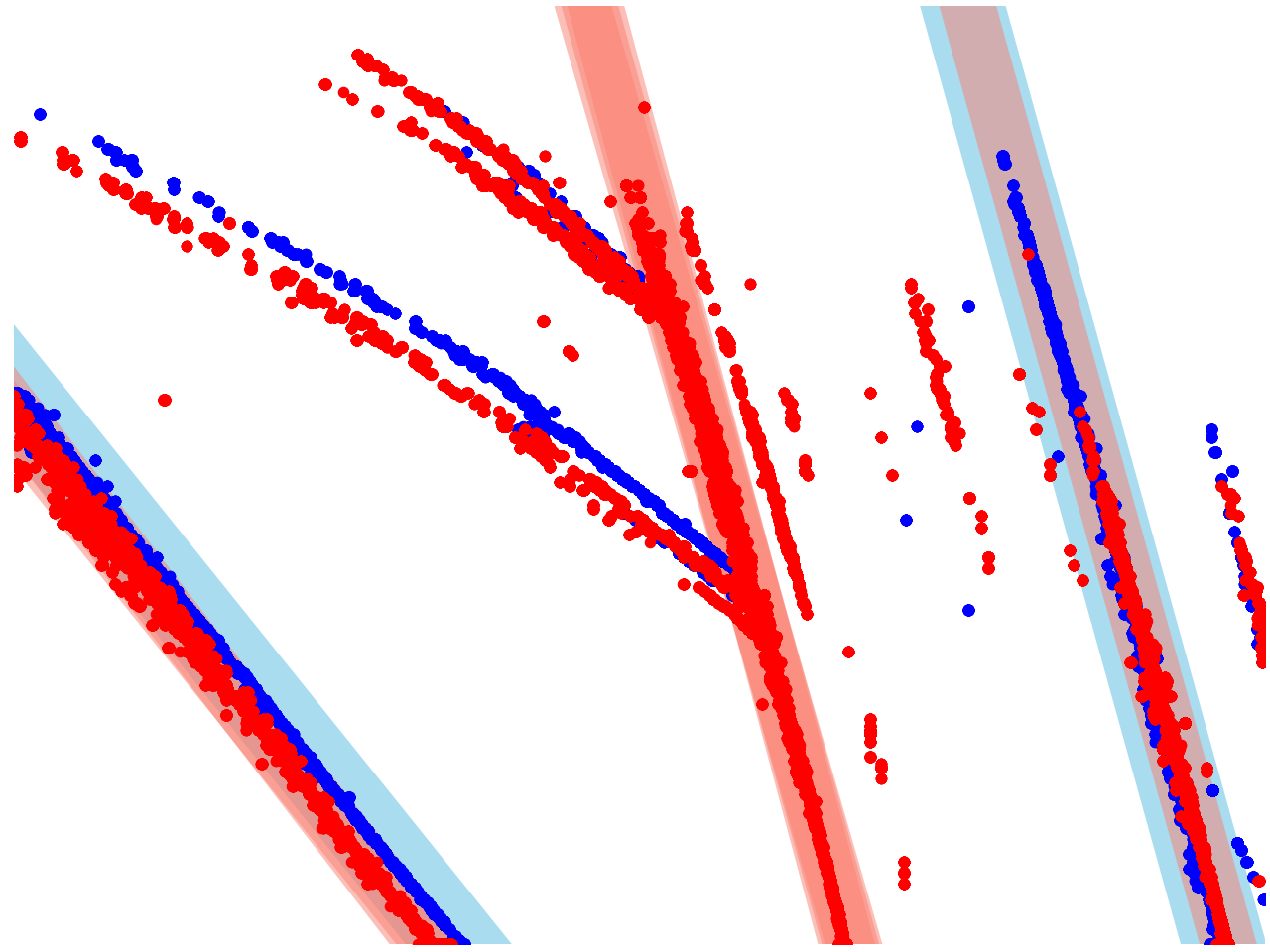}};
        \node (map_image) [process_content,label=above:Map]  at (stop.center) {
        \includegraphics[width=\textwidth, clip]{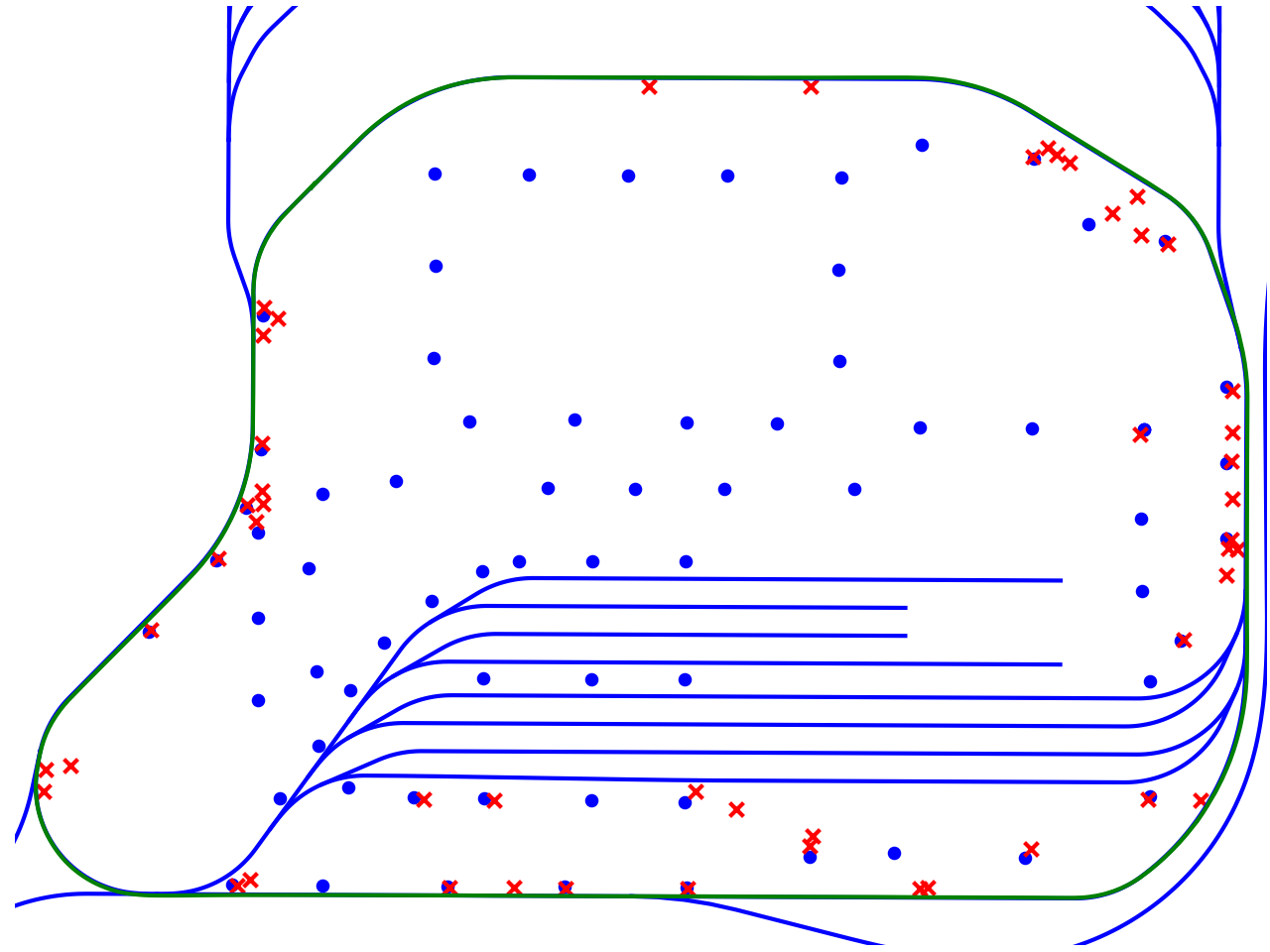}};

        \draw [arrow] (start) -- node [above,midway] {Iterative \acs{HT}} (pro1);
        \draw [arrow] (pro1) -- node[left,midway] {Second \acs{HT}} (pro2);
        \draw [arrow] (pro2) -- node[above,align=center, midway] {Odometry \\
        Triangulation} (stop);
    \end{tikzpicture}

    \caption{Poles are detected and tracked from the event stream using the \finalsubmission{\sout{HoughCeption}Hough$^2$Map} pipeline. 
    Subsequently, in combination with an odometry these poles can be triangulated and placed on a map.}
    \label{fig:full_pipeline}
    \vspace{-0.6cm}
\end{figure}

Otengui~\etal~\cite{Otegui2017} summarized many works to achieve on-board positioning, mainly through a combination of \ac{GNSS} and \ac{IMU} measurements.
However, to achieve high levels of safety and reliability, estimates from independent sensor modalities should be combined~\cite{Beugin2012Simulation-basedLocalization}.
Tschopp~\etal~\cite{Tschopp2019ExperimentalVehicles} and Burschka~\etal~\cite{Burschka2020OpticalEnvironments} demonstrated the applicability of using visual-aided odometry on railways. This modality has shown great success in other domains such as \acp{MAV}~\cite{Burri2015Real-timeEnvironments, Fehr2018Visual-InertialTango}, autonomous driving~\cite{Burki2019AppearancebasedLocalization} and handheld augmented/virtual reality~\cite{Arth2011Real-timeDevices,Lynen2015, Lynen2015GetLocalization}, but has not yet fully been transferred to the domain of rail vehicles.

\reviewchanges{A typical (visual) localization system consists of a means to estimate relative motions (e.g. odometry), the detection and mapping of relevant landmarks and the association of such landmarks to a previously built map or ground truth map in order to reset drift accumulated from the odometry module~\cite{Fehr2018Visual-InertialTango}.}

In this work, \reviewchanges{we focus on one part of such a system, namely the detection and mapping of relevant landmarks by introducing \finalsubmission{\sout{HoughCeption}Hough$^2$Map} as outlined in Figure~\ref{fig:full_pipeline}. \sout{we propose a novel mapping solution for relevant landmarks in the environment as a part of a visual-aided global localization pipeline.}}
In particular, detecting, tracking and mapping of meaningful infrastructure elements, such as the poles of electrical power lines, signal or phone lines, proves to be beneficial, as they can be referenced to accurate manually surveyed ground truth maps for localization.
Additionally, all these infrastructure elements can be predominantly characterised and visually detected as vertical lines.

In order to tackle typical challenges in rail environments, such as motion blur resulting from high vehicle speeds and difficult lighting conditions~\cite{Tschopp2019ExperimentalVehicles}, we propose the use of an event-based camera, also known as a \ac{DVS}.
These can achieve significantly higher temporal resolutions and dynamic range compared to conventional cameras~\cite{Mahowald1992,Lichtsteiner2008, Brandli2014} making them an ideal candidate for use in railway scenarios.
Compared to conventional cameras, which periodically output intensity values of entire image frames, \acp{DVS} produce asynchronous event tuples $e = \langle x,y,t,p \rangle$, where $(x,y)$ are the pixel positions on the image sensor, $t$ the corresponding time-stamps and $p$ the polarities, indicating whether the independent pixel intensities have increased or decreased \reviewchanges{more than a certain threshold}. 

\reviewchanges{
\begin{figure} 
    \centering
      \subfloat[]{%
        \includegraphics[width=0.45\linewidth]{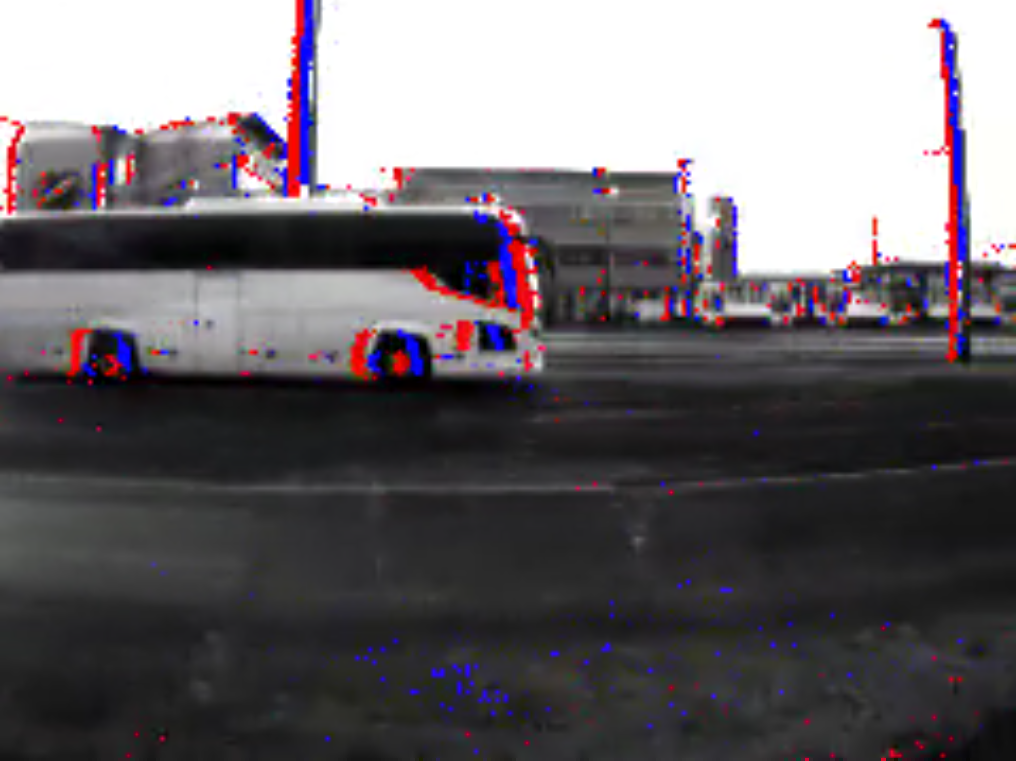}
        }
        \hfill
          \subfloat[]{%
      \includegraphics[width=0.45\linewidth]{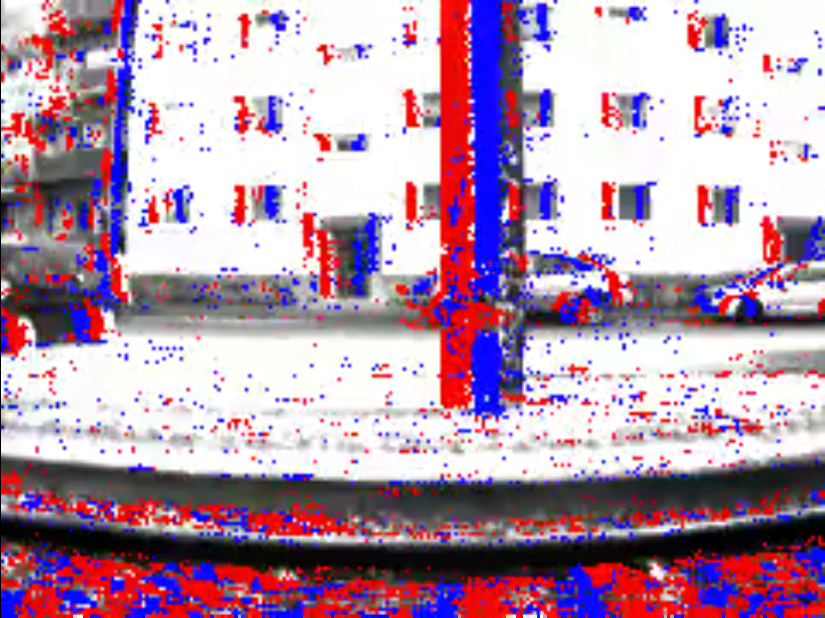}
    }
  \caption[Example of pole observations on \texttt{Trajectory 1} (left) and \texttt{Trajectory 2} (right).]{\reviewchanges{
  Example of pole observations on \texttt{Trajectory 1} (left) and \texttt{Trajectory 2} (right). For more information about the trajectories, refer to Section~\ref{sec:datasets}. For visualization purposes, the events are accumulated for one event \textit{array}\footnotemark (which leads to a motion blur effect in the visualization) and overlaid on a conventional camera image. Illumination increases (positive events) are represented in blue while illumination decreases (negative events) are represented in red.}
  }
  \label{fig:examples} 
  \vspace{-0.6cm}
\end{figure}
\footnotetext{Event arrays are bundles of events from the last $\unit[33]{ms}$~\cite{Brandli2014}.}
Figure~\ref{fig:examples} showcases how typical poles are observed as vertical structures by the \ac{DVS} for both investigated datasets (see Section~\ref{sec:datasets}).
More examples can be found in our paper video\footnote{Video available at \url{https://youtu.be/YPSiODVzD-I}}.
}
Due to \reviewchanges{\sout{this}the} unique data stream, novel algorithms need to be devised to achieve real-time performance~\cite{Benosman2012, Benosman2014,Mueggler2017}.

Hence, in this work, lines from infrastructure landmarks are detected and tracked in real-time using a concatenation of custom, event-based \acp{HT} and further triangulated based on a given odometry.
Concisely, the contributions of this work are the following:
\begin{itemize}
\reviewchanges{
\item A novel real-time, iterative, event-based \ac{HT} and \ac{NMS} strategy to detect straight edges produced by infrastructure, even at close distances to the vehicle and at high velocities.
\item A full real-time mapping pipeline to detect, track and triangulate meaningful landmarks characterized by vertical lines, given an available odometry.
}
\item An evaluation of mapping accuracy and reliability of the whole pipeline on real-world data collected on a tram in Potsdam, Germany. 
This includes a quantitative evaluation on a shorter track with a high-accuracy surveyed ground truth map and a qualitative evaluation on a longer track segment. 
\item An in-depth analysis of additional and missed detections.
\end{itemize}
%

\section{Related Work}
\label{sec:rel_work}
\textbf{Railway mapping:}
To enable train operation which is not limited by the infrastructure-side Balises, current research focuses on on-board localization~\cite{Otegui2017, Siebler2020}.
Leveraging the unique constraint that rail vehicles can only operate on their tracks, multiple works investigated the use of digital maps of the rail network to improve localization accuracy using \ac{GNSS}~\cite{Gerlach2009b,Hasberg2012,Heirich2013a}, \reviewchanges{odometry map matching~\cite{Winter2020Train-borneEvaluation} which can be problematic on parallel tracks or long segments,} eddy current sensors~\cite{Bohringer2006a,Hensel2011b} or \ac{IMU} vibration patterns~\cite{Heirich2013}.
However, the surroundings of the tracks are often not well mapped even though it can be highly informative.
Visual mapping using conventional cameras was demonstrated by Wohlfeil~\etal~\cite{Wohlfeil2011}, mapping both tracks and switches allowing for global localizations close to the switches. 
\finalsubmission{\sout{HoughCeption}Hough$^2$Map}, in contrast, focuses on mapping the more frequent infrastructure elements such as poles, especially visible from a side-facing camera.

\textbf{Line detection:}
In order to map such infrastructure elements, vertical lines in the environment need to be detected.
Hough~\cite{Hough1962} introduced an approach for detecting simple shapes like lines in images based on a voting scheme within a pre-defined parameter space.
Improvements to the \ac{HT} over the years include improved line parameterization to better account for vertical lines~\cite{Duda1972}
and performance improvements by introducing adaptive parameter spaces of varying resolutions and searching through them in a hierarchical manner~\cite{Kittler1987,Li1986} or probabilistic sub-sampling of voting points~\cite{Matas2000RobustTransform}. 
\reviewchanges{By observing the development of peaks in the \ac{HS} in an incremental probabilistic \ac{HT}, early stopping can be performed~\cite{YlaJaaski1994}.
Dalitz~\etal~\cite{Dalitz2017} replaced the \ac{NMS} with an iterative scheme consisting of a simple maximum search and the subsequent removal of all points from the \ac{HS} belonging to the current maximum.
}

%

A second class of line detectors are based on non-parametric detection~\cite{Etemadi1992RobustData,Burns1986ExtractingLines,Kahn1990ANavigation}.
Improved real-time versions, such as \ac{LSD} by Grompone von Gioi~\etal~\cite{GromponeVonGioi2010LSD:Control} and ELSD by Patraucean~\etal~\cite{Patraucean2012AFitting}, allow for fast and accurate detection with minimal false positives.

All aforementioned methods operate on standard gray-scale images.
However, to leverage the benefits of event-based vision, most of those algorithms need to be revised and adapted.
That is one of the main goals of this paper.

\textbf{Event-based line detection:}
Conradt~\etal\cite{Conradt2009} introduced an event-based \ac{HT} for balancing a pencil, modeling the detected line as a Gaussian in the \ac{HS} which gets updated by incoming events, allowing for fast updates but limiting the system to just one tracked line. 
Ni~\etal~\cite{Ni2012} used a circular \ac{HT} to track microparticles, reducing the computational load by limiting the parameter space to a fixed specific radius. 
Glover~\etal~\cite{Glover2016} introduced a circular \ac{HT} for detecting and tracking balls in cluttered environments.
To reduce the search space the \ac{HT} is directed, only permitting events which, according to their respective local optical flow, agree with the motion model of the tracked object.
The significant problem is the search for the maximum in the \ac{HS}, which can be solved by the usage of \acp{SNN}~\cite{Seifozzakerini2016}.
Yuan and Ramalingam~\cite{Yuan2016} developed an event-based pose estimation algorithm based on vertical lines in the environment which are detected using a binning scheme similar to a simple \ac{HT} with a 1D parameter space, thus significantly reducing the possibilities of detectable lines, but enabling real-time performance. 
Mueggler~\etal~\cite{Mueggler2014} introduced a similar event-based pose estimation but for 6 \ac{DoF} using a black square as reference in the event stream.
Line detection is performed by updating known lines with events if these are significantly close to the known line. Only the first initialization is performed using a full \ac{HT}. 
\reviewchanges{Bagchi and Chin~\cite{Chin2019} perform attitude estimation by tracking stars as straight lines in a spatio-temporal space, instead of detecting actual lines in the image space.
The estimation is done iteratively per event, but only inside a fixed time window and lacks a decay of old events, it does therefore not continuously track the stars.
}

As an alternative to a \ac{HT}, Brandli~\etal~\cite{Brandli2016} introduced an event-based line segment detector similar to the non-parametric\reviewchanges{\sout{ classical}} methods\reviewchanges{ mentioned above}.
Le Gentil and Tschopp~\etal~\cite{Gentil2020IDOL:Lines} detect line segments as locally spatio-temporal planar patches to perform visual-inertial odometry.
Both approaches, while promising, are not able to function in real-time in scenarios with high apparent motion, e.g. texture-rich scenes with fast motion.
\reviewchanges{Valeiras~\etal~\cite{ReverterValeiras2019} perform event-based line detection by modelling lines in a spatio-temporal space and matching events to line hypothesis using an optical flow for events. In comparison with our approach, they have to limit the maximum number of hypotheses, and have a higher parametrization complexity.
}

\section{\finalsubmission{\sout{HoughCeption}Hough$^2$Map}}
\label{sec:method}
In this section, the whole \finalsubmission{\sout{HoughCeption}Hough$^2$Map} pipeline, as shown in Figure~\ref{fig:full_pipeline}, is introduced.
First we explain how vertical structures are efficiently detected using an iterative event-based \ac{HT} and \ac{NMS} in comparison to a classical full \ac{HT}.
Using a second spatio-temporal \ac{HT} the structures are tracked over time and subsequently triangulated in combination with an available odometry to obtain their position in the map. 

\subsection{Iterative Hough Transform -- Pole Detection}
\label{sec:iht}
\ac{HT}~\cite{Hough1962} is a classic method for detecting lines in images.
Given that each line can be represented by a distance $r$ \reviewchanges{from the top left corner} and an angle $\theta$, the \reviewchanges{\sout{\ac{HS}}\acf{HS}} represents the parameter space of possible lines that are being detected, where each point in the \finalsubmission{\sout{image space}\ac{HS}} corresponds to a line hypothesis.
More precisely for a point at image coordinates $\left(u,v\right)$ the added hypotheses in \reviewchanges{the} \ac{HS} are all points $p_h$,
\begin{equation} \label{eq:line}
p_h \in \{(r, \theta)\, |\, r = u\cdot\cos{\theta} + v\cdot\sin{\theta}, \theta \in [\,0, \pi]\, \}.
\end{equation}
For practical reasons the \ac{HS} is \reviewchanges{\sout{quantized}discretized} into $N$ and $M$ intervals for distances and angles, respectively.
\reviewchanges{Thereby, to detect mainly vertical structures, we restricted $\theta \in [\, \unit[-10]{^\circ}, \unit[10]{^\circ}]$.}
After accumulating hypotheses from all points in the image space and adjusting for camera distortions using a pre-computed look-up table, distinct lines in the image space will appear as local maxima in the \ac{HS}.

To reduce the amount of noise, a \ac{NMS} procedure is applied to determine the valid maxima in the \ac{HS}.
First, we define the set of local maxima $\mathcal{M}_l^t$ at timestep $t$ as all points that are above a threshold $T_{\text{Hough}}$ and are larger than all their neighbours in an 8-connected neighbourhood.
Additionally, the set of global maxima $\mathcal{M}_g^t \subseteq \mathcal{M}_l^t$ includes all local maxima that \finalsubmission{are bigger than all other local maxima} within a defined suppression radius $R$\finalsubmission{\sout{ are bigger than all other local maxima}}.
The objective of \ac{NMS} is to find $\mathcal{M}_g^t$ inside the \ac{HS}, which represents a set of most distinct lines in the image.
To perform \ac{NMS}, all $\mathcal{M}_l^t$ in the \ac{HS} are detected and ordered.
Afterwards, the ordered $\mathcal{M}_l^t$ are successively evaluated and marked as belonging to $\mathcal{M}_g^t$ if there does not already exist\finalsubmission{\sout{s}} another global maximum within the suppression radius.

Detecting lines in a \ac{DVS} event stream can be achieved by accumulating the events inside a sliding window, applying \ac{HT} as explained before to this window and retrieving $\mathcal{M}_g^t$ in the \ac{HS}.
\reviewchanges{The sliding window is defined as the last $300$ events which were observed by the \ac{DVS}.}
When a new event arrives it can be added to the window, the oldest event removed and the \reviewchanges{\sout{procedure}line detection} can be repeated.
A computationally efficient approach is to update the \reviewchanges{accumulator cells of the} \ac{HS} iteratively \reviewchanges{per event} by only adding the hypotheses of the newest point and similarly removing the ones of the oldest point\reviewchanges{. This results in two sets of accumulator cells, $\mathcal{P}_+$ and $\mathcal{P}_-$, which need to be updated for the added and removed event, respectively}.
The major disadvantage is that at each step a full traversal of the \ac{HS} is still required to perform \ac{NMS}, which makes this approach computationally inefficient and prevents real-time operation on a \ac{DVS}.
To tackle this, we developed an iterative \ac{NMS} for event-streams in Section~\ref{sec:nms}.

\subsection{Iterative Non-Maxima Suppression}
\label{sec:nms}
To perform \ac{NMS} more efficiently we have to observe that each added or removed event to the iterative \ac{HT} explained in Section~\ref{sec:iht} modifies at most $\reviewchanges{O(}\max(N, M)\reviewchanges{)}$ \reviewchanges{\sout{hypotheses}accumulator cells} in the \ac{HS}.
Therefore, the changes in the \ac{HS} per event are minimal.
By only examining the \reviewchanges{two} set\reviewchanges{s} of affected \reviewchanges{\sout{points}accumulator cells} $(\mathcal{P}_+,\mathcal{P}_-)$ \reviewchanges{\sout{where the \ac{HS} was incremented or decremented}that were increased or decreased  by the new or removed event}, respectively, we can update the list of global maxima from the previous step $\mathcal{M}_g^{t-1}$ \reviewchanges{to obtain $\mathcal{M}_g^t$}, without having to iterate over the entire \ac{HS}.
A detailed description of \reviewchanges{\sout{an}the proposed} iterative \ac{NMS} is depicted in Algorithm~\ref{alg:nms}.
This implementation of an iterative \ac{NMS} leads to the exact same results as performing a full \ac{NMS} for each event, as it considers all possible ways in which $\mathcal{P}_+$ and $\mathcal{P}_-$ can influence the \ac{HS} to cause new global maxima to appear or disappear \reviewchanges{\sout{from}with respect to} $\mathcal{M}_g^{t-1}$.

\begin{algorithm}[t]
\begin{algorithmic}[1]
\Procedure{INMS}{$\mathcal{P}_+^t,\mathcal{P}_-^t, \mathcal{M}_g^{t-1}$}
\Algphase{Phase 1 - Obtain candidates for $\mathcal{M}_g^{t}$}
\State $\mathcal{M}_g^{t*} \leftarrow \mathcal{M}_g^{t-1}$\Comment{Init list of candidate global maxima}

\ForAll{$p_+\in \mathcal{P}_+^t$}
\State skip $\leftarrow$ \False

\ForAll{8-connected neighbours $p'_+$ of $p_+$}
\If{HS$(p_+)$ is equal to HS$(p'_+)$}
\State skip $\leftarrow$ \True
\If{$p'_+\in\mathcal{M}_g^{t-1}$}
\State Remove $p'_+$ from $\mathcal{M}_g^{t*}$
\State Add $\left\{m_l^t \in \mathcal{M}_l^t \,\big|\, ||m_l^t-p'_+ ||\leq R\right\}$ to $\mathcal{M}_g^{t*}$
\EndIf
\EndIf
\EndFor

\If{\reviewchanges{\textbf{not} skip}}
\If{$p_+\in\mathcal{M}_l^t$ and $p_+\notin\mathcal{M}_g^{t-1}$}
\State Add $p_+$ to $\mathcal{M}_g^{t*}$
\EndIf
\EndIf
\EndFor

\ForAll{$p_-\in\mathcal{P}_-^t$ }

\ForAll{8-connected neighbours $p'_-$ of $p_-$}
\If{HS$(p_-)+1$ is equal to HS$(p'_-)$}
\If{$p'_-\in{\mathcal{M}_l^t}$}
\State Add $p'_-$ to $\mathcal{M}_g^{t*}$
\EndIf
\EndIf
\EndFor

\If{$p_-\in\mathcal{M}_g^{t-1}$}
\State Remove $p_-$ from $\mathcal{M}_g^{t*}$
\State Add $\left\{m_l^t \in \mathcal{M}_l^t \,\big|\, ||m_l^t-p_- ||\leq R\right\}$ to $\mathcal{M}_g^{t*}$
\EndIf

\EndFor

\Algphase{Phase 2 - Detect and apply suppression}

\Repeat
\State suppress $\leftarrow$ \False
\State Sort $\mathcal{M}_g^{t*}$ in descending order

\ForAll{${}^im^t_g\in\mathcal{M}_g^{t*}$}
\State $\mathcal{M}_s^t \leftarrow \left\{{}^jm_l^t \in \mathcal{M}_l^t \,\big|\, i \neq j,\, ||{}^jm_g^t - {}^im_g^t || \leq R \right\}$
\State Remove $\mathcal{M}_s^t$ from $\mathcal{M}_g^{t*}$

\ForAll{$m_s^t\in\mathcal{M}_s^t$}
\ForAll{$m_g^{t-1}\in\mathcal{M}_g^{t-1}$}

\If{$m_s^t = m_g^{t-1}$}
\State Add $\left\{m_l^t \in \mathcal{M}_l^t \,\big|\, ||m_l^t-m_s^t||\leq R\right\}$ to $\mathcal{M}_g^{t*}$
\State suppress $\leftarrow$ \True
\EndIf
\EndFor
\EndFor
\EndFor
\Until{suppress is \False}
\State $\mathcal{M}_g^t \leftarrow \mathcal{M}_g^{t*}$
\State \Return $\mathcal{M}_g^t$
\EndProcedure
\end{algorithmic}
\caption{Iterative Non-Maxima Suppression}
\label{alg:nms}
\end{algorithm}

The algorithm can be divided into roughly two stages.
In the first step, based on $\mathcal{P}_+$ and $\mathcal{P}_-$ as well as $\mathcal{M}_g^{t-1}$, we create a list of potential global maxima \reviewchanges{$\mathcal{M}_g^{t*}$}, without iterating over the entire \ac{HS}\reviewchanges{\sout{ to find $\mathcal{M}_l^t$}}.
\reviewchanges{\sout{Instead, $\mathcal{M}_l^t$ is only partially calculated by checking the 8-connected neighborhood of only certain points of interest within the suppression radius $R$, as depicted in Algorithm~\ref{alg:nms}.}}
\reviewchanges{This is done by checking four distinct cases, out of which two deal with the direct creation of new potential global maxima.
If an accumulator cell in $\mathcal{P}_+$ is a local maximum, now it is also a new potential global maximum, and similarly if an accumulator cell in the 8-connected neighbourhood of a cell in $\mathcal{P}_+$ just became a new local maxima at time step $t$, it is also a potential new global maxima.
The other two cases deal with the potential removal of old global maxima, in which case a more exhaustive search in the suppression radius is required to retrieve the current local maxima that might now have become global maxima.}
In the second step, we \reviewchanges{sort the candidate maxima $\mathcal{M}_g^{t*}$ and filter out the maxima that are within each others\sout{repeatedly apply the}} suppression radius\reviewchanges{.
The removal of a maxima candidate that was a global maxima at time step $t-1$ requires the addition of the local maxima surrounding the removed point to the set of candidates $\mathcal{M}_g^{t*}$.
In this case, the entire second step needs to be repeated, to ensure that none of the new points are within existing suppression radii.\sout{to the list of potential global maxima to obtain $\mathcal{M}_g^t$}
After the repeated filtering and suppression step, the remaining candidate maxima are the new global maxima $\mathcal{M}_g^t$}.

Performing a full \ac{NMS} has a complexity of $O(NM)$, since a full traversal of the \ac{HS} is required.
In contrast the iterative \ac{NMS} has a varying cost, depending on the number of maxima that change during one step.
In the best case scenario, where no changes occur, the complexity is $O(\max(N,M))$, while in the worst case scenario the whole \ac{HS} will be searched, leading to a complexity of $O(NM)$.
In practice though the worst case scenario rarely happens and the iterative \ac{NMS} is far more efficient than the full \ac{NMS} (see Section~\ref{sec:timings-iterative-nms}).

\subsection{Second Hough Transform -- Pole Tracking}
\label{sec:2ndht}
\begin{figure}[t]
\centering
\includegraphics[width=\columnwidth, trim={0cm 0cm 0cm 0cm}, clip]{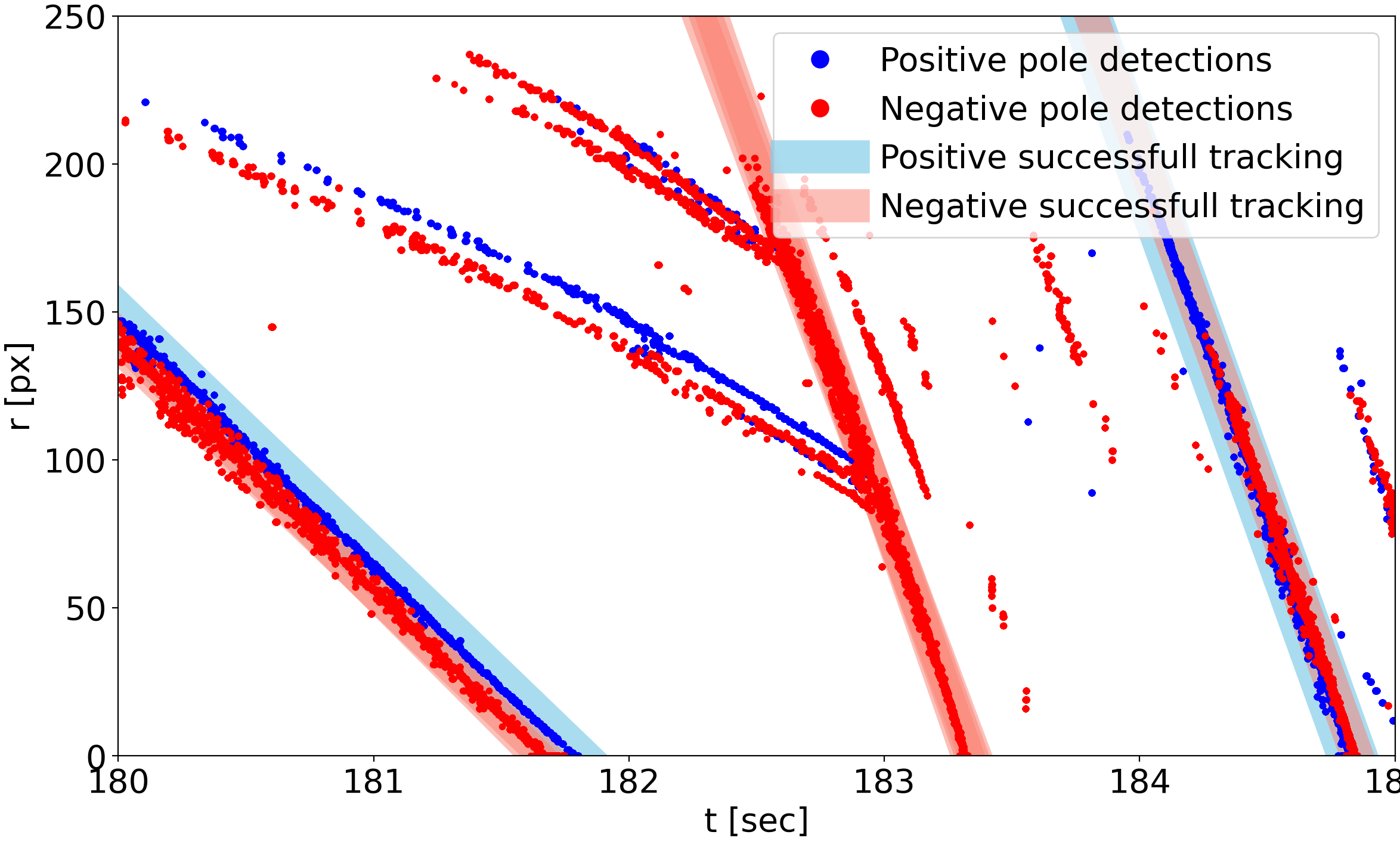}
\caption{The second \ac{HS} tracks the horizontal position of line detections over time for both positive and negative events separately. \reviewchanges{By using the second \ac{HT} on this spatio-temporal representation, object tracks of corresponding detections can be obtained.} The object is only accepted if positive and negative tracks are pairwise, e.g. the middle negative track is discarded.}
\label{fig:second_stage_x_t_space}
\vspace{-0.6cm}
\end{figure}

The next step of the pipeline consists of associating line detections from the same objects across time, in order to triangulate them into map points.
\reviewchanges{We map individual line detections into a spatio-temporal space, by taking the horizontal line positions, as defined by their shortest distance $r$ to the origin (Equation~\ref{eq:line}), at each time of detection, as shown in Figure~\ref{fig:second_stage_x_t_space}).
Assuming a locally constant train velocity, straight lines in the spatio-temporal space correspond to vertical structures which move at constant speed through the \ac{FoV} of the camera.
\sout{When transforming individual line detections into a spatio-temporal space (horizontal line positions,
over time shown in Figure~\ref{fig:second_stage_x_t_space}) and assuming a locally constant train velocity, straight lines can be observed which correspond to vertical structures which move at constant speed through the \ac{FoV} of the camera.}}
This constant velocity assumption only needs to hold true \reviewchanges{for a very short time, \sout{for}} the time a pole is in the cameras \ac{FoV}\reviewchanges{,} and is justified by \reviewchanges{the} high inertias in typical railway vehicles.
A second \reviewchanges{\sout{spatio-temporal}}\ac{HT} can be applied \reviewchanges{on the spatio-temporal representation depicted in Figure~\ref{fig:second_stage_x_t_space}} to associate \reviewchanges{\sout{the }}individual detections with landmarks \reviewchanges{and obtain landmark tracks $D_{t:t+k}$ containing horizontal line position and timestamp pairs, where $k$ depicts the length of the track}. 
\reviewchanges{\sout{The resulting lines directly contain the full track information $D_{t:t+k}$ where $k$ depicts the length of the track.}}
Additionally, we perform the tracking for both the positive and negative events in separate \acp{HS} in order to filter detections triggered by positive events which do not have a corresponding detection triggered from negative events or vice versa, such as the second detection in Figure~\ref{fig:second_stage_x_t_space}.
This filters out larger objects such as buildings or bridges, where the positive and negative line detections occur far apart and which are currently out of the scope of this work.

\reviewchanges{
\subsection{Triangulation}
}
In order to triangulate the tracked detections into landmarks\reviewchanges{, we leverage position estimates $P_{t:t+k} \in SE(2)$ of the vehicle from an external odometry. \sout{based on~[46] we utilize an external odometry.}}
Odometries with locally low drift exist based on Doppler radars, wheel encoders~\cite{Palmer2018} or also front facing cameras~\cite{Tschopp2019ExperimentalVehicles,Burschka2020OpticalEnvironments,Tschopp2020,Bah2009}.
%
%
%
\reviewchanges{
Based on a camera calibration, namely the horizontal principal point $u_0$ and focal length in pixels $\alpha_x$, and samples $D_{t:t+k}$ along the detection track, the landmark position $X$ is obtained using a 2D version of a \ac{DLT} triangulation by assembling a matrix $A \in \mathbb{R}^{k \times 3}$:
%
%
%
%
\begin{equation}
    A[i,j] = (D_{t+i} - u_0)/\alpha_x \cdot P_{t+i}[1,j] - P_{t+i}[0,j],\\
\end{equation}
where $i, j$ are matrix indices.
After computing the \ac{SVD} of A, i.e. $U \cdot \Sigma \cdot V^* = \svd(A)$, we take the least significant right-singular vector of $V^*$, which corresponds to the homogeneous representation of $X$ and just needs to be normalized~\cite{Hartley}.
}
%

    

\section{Experimental Evaluation}
\label{sec:experiments}
To evaluate the accuracy and performance of \finalsubmission{\sout{HoughCeption}Hough$^2$Map}, we performed multiple experiments on real-world railway datasets. Furthermore, an evaluation is performed to compare the iterative \ac{HT} and \ac{NMS} to their conventional counterparts.
\reviewchanges{Although a comparison with ~\cite{ReverterValeiras2019} would be informative, at the time of this publication the authors had not made neither the data nor implementation publicly available.}

\subsection{Datasets}
\label{sec:datasets}
The pipeline was evaluated on two available\reviewchanges{\sout{ pre-recorded}} datasets from a tram in Potsdam, Germany~\cite{Tschopp2019ExperimentalVehicles}. 
\texttt{Trajectory~1} is recorded on the Betriebshof, a short $\unit[815]{m}$ circular track with a large variety of visual features and challenging situations including significant velocity changes.
For this track a high accuracy manually surveyed map of nearly all poles exists which will be used as ground truth for the evaluation of the \reviewchanges{accuracy of the pipeline}. 
\texttt{Trajectory~2} is a longer $\unit[2545]{m}$ section of a public tram line passing through rural, \mbox{(sub-)urban} areas and forests with a more constant vehicle velocity and is therefore a more realistic scenario for everyday railway operation.
Unfortunately, for \texttt{Trajectory~2} no accurate ground truth map was available.
To perform a qualitative evaluation pole positions were manually annotated from high-resolution satellite imagery and aligned to \ac{GNSS} coordinates.

\subsection{Hardware Setup}

\subsubsection{Devices}
\begin{figure}[t]
 \centering
 \includegraphics[width=\columnwidth]{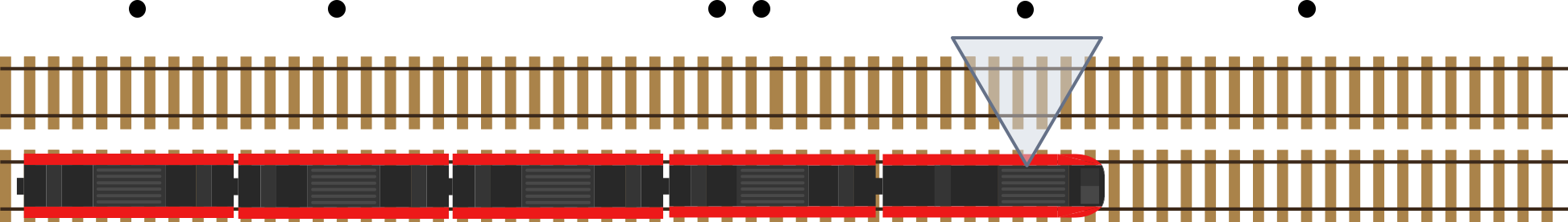}
 \caption{The \ac{DVS} in a perpendicular setup to the \reviewchanges{\sout{trains }}direction of motion \reviewchanges{of the train}, observing the environment passing by.}
 \label{fig:dvs_train_setup}
 \vspace{-0.1cm}
\end{figure}

The dataset was recorded using an iniVation \ac{DAVIS} 240\reviewchanges{~\cite{DavisData2019} with a sensor size of $\unit[240\times180]{px}$, a minimum latency of $\unit[12]{\mu s}$ and a horizontal \ac{FoV} of $\unit[56]{^\circ}$. The device was} oriented perpendicular to the\reviewchanges{\sout{ tram's}} axis of motion \reviewchanges{of the tram} out of a side window as shown in Figure~\ref{fig:dvs_train_setup}. 
This enables a good observability of the environment close to the track and results in a linear perceived motion of detections through the \ac{FoV} enabling easy tracking.
%
As the goal of this evaluation is to assess the accuracy of the detection and mapping pipeline, a high-accuracy \ac{INS} is used as odometry as opposed to drifting odometries to avoid external error sources.
%
%

\subsubsection{Calibration}
The intrinsic camera calibration parameters of the \ac{DAVIS} were obtained using the \textit{Kalibr} toolbox~\cite{Furgale2013UnifiedSystems} on the active pixels.
At the time of the recording of the dataset no appropriate calibration procedure for the extrinsic calibration of the camera relative to the train was available. 
This was therefore obtained through manual tuning of the lateral and longitudinal offset along the train.

\subsection{Iterative Hough Transform and Non-Maxima Supression}
\label{sec:timings-iterative-nms}
To mainly detect vertical lines in the full image, the \ac{HS} was chosen as $r\in\left[\unit[0]{px},\unit[260]{px}\right]$ and $\theta \in \left[\unit[-10]{^\circ}, \unit[10]{^\circ} \right]$ \reviewchanges{in $\unit[1]{^\circ}$ increments}.
In order to compare the performance of our iterative \ac{HT} and \ac{NMS} against the conventional version with a Full-\ac{NMS}, we compared the resulting output of the detection stage (without tracking and mapping) on the \texttt{Trajectory~1} dataset.
A comparison of each individual detection showed that both variants provide the exact same output.
On average the processing of an event in the full \ac{HT} took $\unit[6.04]{\mu s}$ per event $e$ with a maximum of $\unitfrac[30.27]{\mu s}{e}$. In comparison, our iterative variant took on average $\unitfrac[0.65]{\mu s}{e}$ with a maximum of $\unitfrac[6.20]{\mu s}{e}$.
\reviewchanges{For event rate statistics, see Section~\ref{sec:timings}.}
%

\subsection{Detection Reliability and Accuracy}

\begin{figure}[t]
\centering
\includegraphics[width=0.9\columnwidth,  clip]{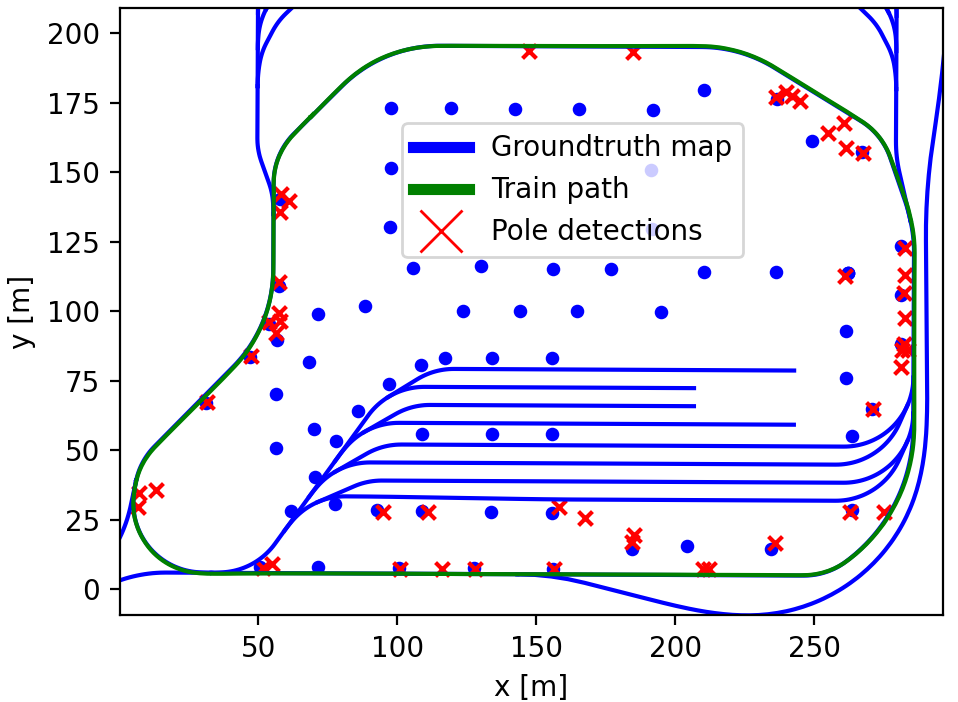}
\caption{Infrastructure map of \texttt{Trajectory~1}. Many of the ground truth poles in the center of the loop are occluded from the camera and therefore cannot be mapped. Some false positives are discussed in detail in Section~\ref{sec:results}.}
\label{fig:betriebshof}
\vspace{-0.6cm}
\end{figure}
\begin{figure*}[t]
\centering
\includegraphics[width=0.9\textwidth,trim={0cm 0cm 0cm 0cm}, clip]{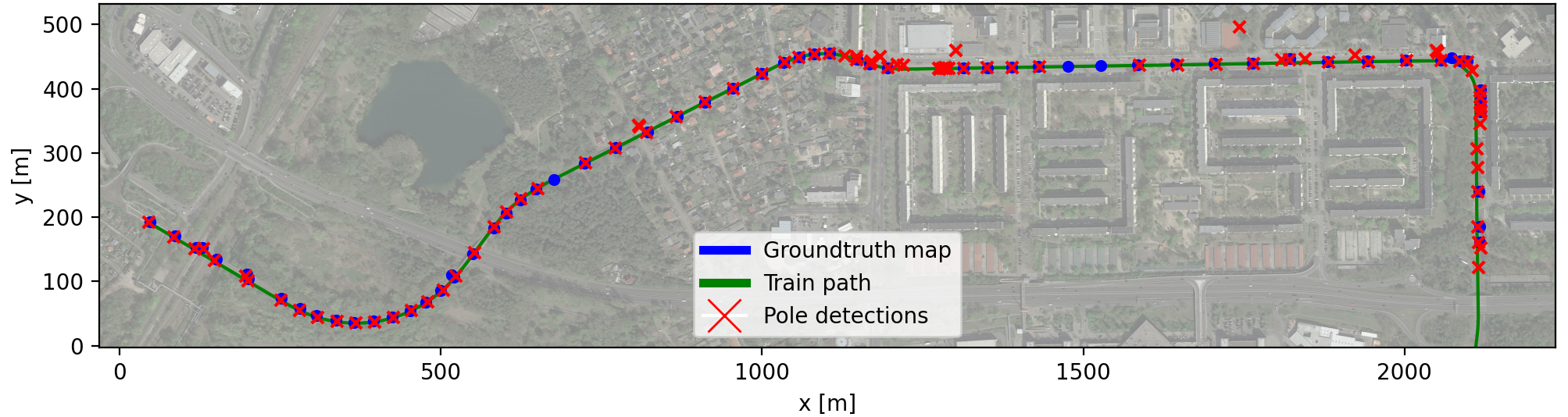}
\caption{Infrastructure map of \texttt{Trajectory~2} overlaid on a satellite image including pole detections and manual annotations. }
\label{fig:full_track_with_detections}
\vspace{-0.3cm}
\end{figure*}
Due to the differences between the datasets as well as the available ground truth data, evaluations are performed differently for each dataset. 
For \texttt{Trajectory~1} an accurate ground truth of pole positions is available.
Therefore, both the reliability of detecting poles or wrongly detecting other objects, as well as the accuracy of the triangulation can be evaluated.
The accuracy is reported as a \ac{RMSE} over all matching detections.
%
%
%
As the manual annotation for \texttt{Trajectory~2} is a fairly inaccurate process, evaluations will be carried out only for the detection reliability and not for the mapping accuracy. 

\label{sec:results}

The full pipeline as described in Section~\ref{sec:method} was tested on both datasets.
All parameters, such as \ac{HS} thresholds, integration window sizes and \ac{NMS} suppression radii were kept constant over both experiments.
Figures~\ref{fig:betriebshof} and \ref{fig:full_track_with_detections} show the resulting maps for the respective datasets including pole detections, \ac{GNSS} odometry and ground truth data.

\subsubsection{Detection reliability}
\label{sec:detection}

\begin{table}[t]
\centering
\caption{Pole detection reliability of the pipeline.}
\begin{tabular}{lcc}
\toprule
        & \texttt{Trajectory~1} & \texttt{Trajectory~2}                 \\  \hline
Ground truth & 43 & 65 \\
True Positives & 23 & 60 \\
False Negatives & 20 & 5 \\
False Positives & 27 & 28\\
\bottomrule
\end{tabular}
\label{table:false_negatives}
\end{table}
The \textit{reliability} evaluation results of the mapping pipeline on both datasets can be found in Table~\ref{table:false_negatives}.
On \texttt{Trajectory~1}, out of 43 known poles in a range of $\unit[25]{m}$ to the tram track 23 were detected successfully. 
In addition 27 other objects were included in the detections as seen in Table~\ref{table:false_postive_detections}. These mostly consisted of objects which are visually similar to poles or \reviewchanges{\sout{parts}infrastructure adjacent to\sout{of}} buildings, such as chimneys \reviewchanges{or rain pipes}, but which were not detailed in the available ground truth data.
%
%
Furthermore, many poles in the ground truth map are quite far away from the tram track, making a reliable detection difficult due to occlusions, \ac{FoV} of the camera, and limited camera resolution. 
\reviewchanges{\sout{This is complicated even further}Further complications are caused} by a strongly varying velocity profile of the tram during this test drive, which is rather uncharacteristic \reviewchanges{\sout{for a }during} normal\reviewchanges{\sout{ tram}} operation.
This leads to \reviewchanges{occasional} violations of the constant velocity assumption specified in Section~\ref{sec:2ndht}. 
These properties make \texttt{Trajectory~1} a rather challenging dataset.

\begin{figure}[ht!]
    \centering

  \subfloat[Changing velocity \label{fig:changing_speed_hough}]{%
      \includegraphics[width=0.45\linewidth]{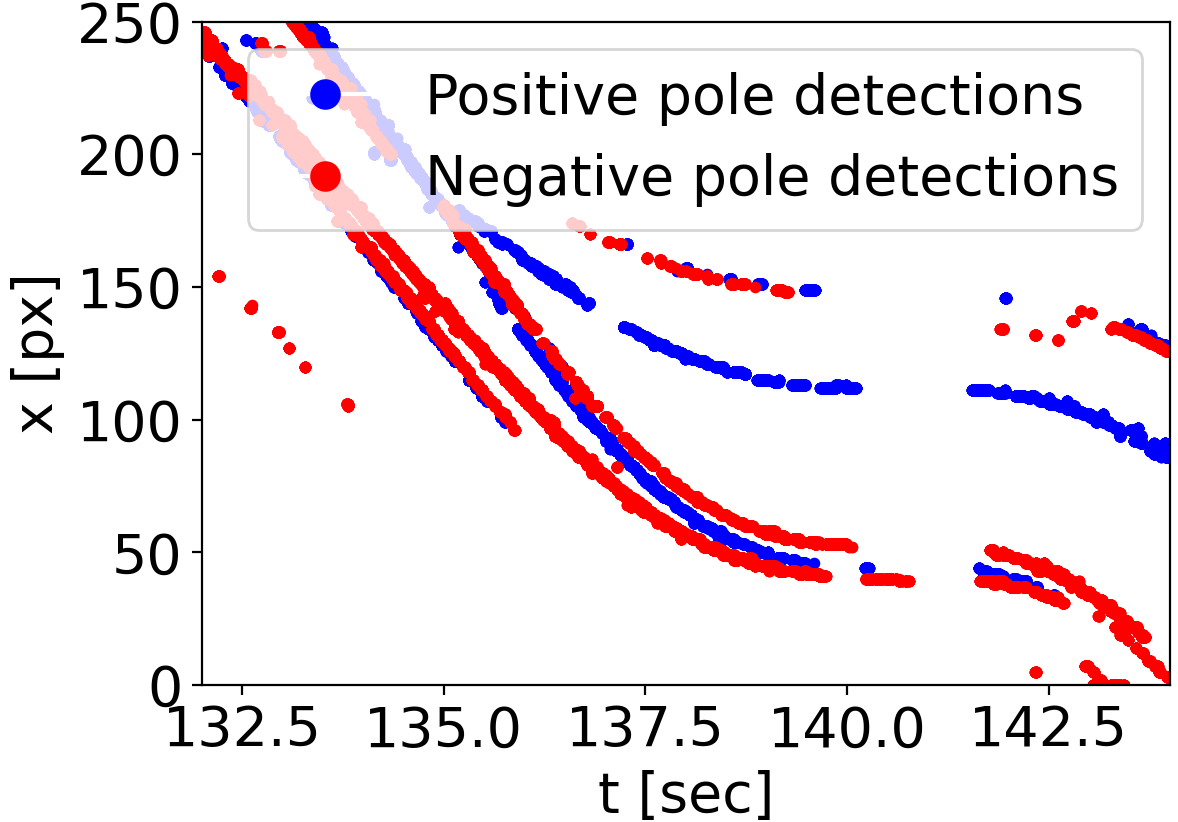}
    }
    \hfill
  \subfloat[Slow speed\label{fig:changin_speed}]{%
        \includegraphics[width=0.45\linewidth]{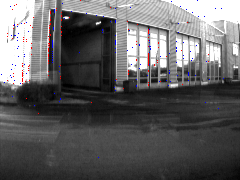}
        }
    \
  \subfloat[Poles occluded by vegetation\label{fig:pole_drowned_in_noise}]{%
        \includegraphics[width=0.45\linewidth]{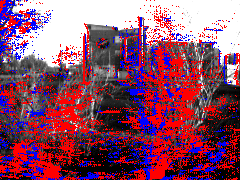}
        }
    \hfill
          \subfloat[Chimney\label{fig:chimney}]{%
      \includegraphics[width=0.45\linewidth]{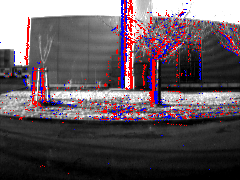}
    }
    \
  \subfloat[Street signs\label{fig:street_signs}]{%
        \includegraphics[width=0.45\linewidth]{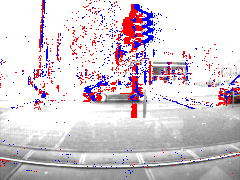}}
    \hfill
  \subfloat[Other tram\label{fig:other_tram}]{%
        \includegraphics[width=0.45\linewidth]{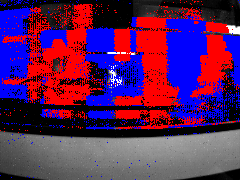}}
    \
  \caption{
  %
  \reviewchanges{(a) shows the spatio-temporal domain in which poles are tracked by the second \ac{HT} (b)-(f) show the \ac{DVS} visualization as in Figure~\ref{fig:examples} to highlight the characteristics of generated events in certain situations and how these might lead to failure cases.}
  (a) and (b) Missed detections due to slow and changing velocity.
  (c) \reviewchanges{\sout{and (d) }}Missed detection\reviewchanges{\sout{s}} due to partial occlusion \reviewchanges{and noise\sout{, noise or size}}.
  (\reviewchanges{\sout{e}d}) Detections of unmapped objects such as a chimney.
  \reviewchanges{\sout{(f) Two poles about to occlude each other.}}
  (\reviewchanges{\sout{g}e}) Street signs and signals wrongly detected as poles.
  (\reviewchanges{\sout{h}f}) Another tram passing by \reviewchanges{with its window edges }causing false detections.
  %
  }
  \label{fig:false_pos_false_neg_t1} 
  \vspace{-0.6cm}
\end{figure}

\begin{table}[t]
\centering
\caption{Distribution and origin of false positive detections}
\begin{tabular}{lcc}
\toprule
 & \texttt{Trajectory~1} & \texttt{Trajectory~2} \\ \hline
Unmapped poles & 8 & 7 \\
\reviewchanges{Adjacent to b}uildings & 17 & 2\\
Trees & 1 & 2\\
Street Signs & 1 & 11\\
Other Train & 0 & 6\\ \hline
Total & 27 & 28\\
\bottomrule
\end{tabular}
\label{table:false_postive_detections}
\vspace{-0.4cm}
\end{table}
%
On \texttt{Trajectory~2} out of 65 known poles 60 were detected successfully. 
There were however also 28 false positive detections from various sources.
As most known poles are closer to the track and as the tram keeps a more steady velocity, this trajectory better agrees with our assumptions of constant velocity and represents a more realistic scenario. 
However, due to the manual annotation process, ground truth data might be incomplete, as several poles were not recognizable due to vegetation or challenging lighting in the satellite image. 
%
%
An in-depth analysis of missed and wrong detections revealed several distinct reasons reported in Figure~\ref{fig:false_pos_false_neg_t1} and Table~\ref{table:false_postive_detections}:
\begin{itemize}
    \item Most commonly, \reviewchanges{manually set }thresholds in the \acp{HS} \reviewchanges{ incorrectly} filter out poles, such as when these are too small, too far away or occluded by surrounding vegetation (Figure~\reviewchanges{\sout{\ref{fig:too_small_pole} and~}}\ref{fig:pole_drowned_in_noise}).
    \item Multiple poles located close to each other \reviewchanges{\sout{(Figure~\ref{fig:occluding_poles}) }}can get filtered by the suppression radius \reviewchanges{\sout{in}of} the \ac{NMS}.
    \item Drastic changes in vehicle speed during a pole observation significantly violate the constant velocity assumption of the second \ac{HT} resulting in a non-straight line for the pole in the $x-t$ space visible in Figure~\ref{fig:changing_speed_hough}.
    \item Very slow speeds result in minimal changes in illumination and a very low event rate mainly dominated by sensor noise \reviewchanges{\sout{visible in}as shown in} Figure~\ref{fig:changin_speed}.
\end{itemize}
Potential measures to improve\reviewchanges{\sout{ on this}} are \reviewchanges{\sout{adaptable}adaptive} thresholds and sensor settings based on the observed environments, or vehicle speed compensation in the $x-t$ space.
Furthermore, if scenarios with highly varying speeds become relevant, a set of \acp{EKF} could be used to track pole detections.
%

Main reasons for additional pole detections include:
\begin{itemize}
    \item 
    %
    From the \ac{DAVIS} video stream a total of 15 poles can additionally be confirmed as correctly detected, but a corresponding ground truth pole is missing as the ground truth data for both trajectories is incomplete.
    \item Street signs or signals  (Figure~\ref{fig:street_signs}) located near the track resemble poles closely resulting in a detection.
    \item At one point in time another tram passes through the \ac{FoV} of the camera (Figure~\ref{fig:other_tram}), causing several wrong detections over a short amount of time. \reviewchanges{Such false positives could be filtered by a minimal distance threshold from the camera, as the additional speed of the other vehicle would result in a very small triangulated depth.}
    \item On rare occasions trees and special features \reviewchanges{\sout{of}adjacent to} buildings, such as chimneys, are also detected (Figure~\ref{fig:chimney}).
\end{itemize}
Many of the false positives correspond to infrastructure objects that can be consistently mapped.
Therefore, for the purpose of railway localization, such objects are not a drawback of our approach but an opportunity to get even more good landmarks into the map.

\subsubsection{Accuracy evaluation}
In a second step we evaluated the \textit{accuracy} of our mapping pipeline.
%
%
Detections were matched to ground truth pole positions through a simple nearest-neighbor scheme with a rejection radius of $\unit[4]{m}$.
From the matching set of detections the \ac{RMSE} can be computed with a value of $\unit[1.1518]{m}$.
This magnitude of error is acceptable for the target application since landmarks can still be distinctively associated for the purpose of localization.
Furthermore, the error consists of a longitudinal shift along the axis of motion with an average of $\unit[1.0045]{m}$ and a lateral shift with an average of $\unit[0.73748]{m}$.
In an \ac{ETCS} context~\cite{LocalizationWorkingGroupLWG2019RailwaysRequirements}, a good localization perpendicular to the axis of motion is of higher importance to facilitate track selectivity.
\reviewchanges{
Even though both reported errors are within the same order of magnitude, the lower error in lateral direction suggests a high potential to enable track selective localization.
Please note however, that the mapping accuracy does not directly correspond to a possible pose estimation accuracy but reflects the global error of placing infrastructure elements into the map.
}
%


\subsection{Processing Time} \label{sec:timings}
Table~\ref{table:processing_time_stats} provides an overview of event rate $\epsilon$ statistics, \reviewchanges{namely the number of events per second}, processing times $t_p$ \reviewchanges{of a parallelized version} of the whole pipeline and real time factors on both datasets, \reviewchanges{corresponding to the total CPU load}. 
The size of the parameter space was chosen to be $21\times260$.
All data was processed on an Intel i7-9750H CPU with $\unit[16]{GB}$ of memory.
Thanks to the iterative implementation of \ac{HT} and \ac{NMS}, our pipeline easily achieves real-time performance resulting in an average \reviewchanges{total} CPU load of $\unit[41]{\%}$ and $\unit[59]{\%}$ for \texttt{Trajectory~1} and \texttt{Trajectory~2}, respectively, while still allowing the system to perform other tasks like visual odometry, localization on the same platform or enable real-time processing also for higher vehicle velocities.
\reviewchanges{For a stable control architecture, a constant frequency of the pipeline is desirable.
Running the pipeline also for the cases with very high event rates, for a maximum of $\unit[11.5]{\%}$ of event arrays, the processing of the subsequent event array had to be buffered.
In a delay sensitive application, this can be tackled by randomly sub-sampling events in case of a very high event rate.}
\begin{table}[t]
\centering
\caption{Event rate $\epsilon$ statistics of the received sensor data, processing time $t_p$, as well as a real-time factor describing the required processing time compared to the rate of new sensor measurements \reviewchanges{(total CPU load)}. $\bar{\cdot}$: Average, max: maximum, StD: standard deviation}
\begin{tabular}{lcc}
\toprule
 & \texttt{Trajectory~1} & \texttt{Trajectory~2} \\  \hline
$\bar{\epsilon}$ & $\unitfrac[115633]{e}{s}$ & $\unitfrac[204909]{e}{s}$ \\
$\epsilon_{max}$ & $\unitfrac[824705]{e}{s}$ & $\unitfrac[1197064]{e}{s}$ \\
$\epsilon_{StD}$  & $\unitfrac[133443]{e}{s}$ & $\unitfrac[146519]{e}{s}$\\
\hline
%
$\bar{t_p}$ & $\unitfrac[3.55]{\mu s}{e}$ & $\unitfrac[2.89]{\mu s}{e}$\\
$t_{p,max}$ & $\unitfrac[127.8]{\mu s}{e}$ & $\unitfrac[117.7]{\mu s}{e}$\\
$t_{p,StD}$ & $\unitfrac[5.13]{\mu s}{e}$ & $\unitfrac[3.47]{\mu s}{e}$ \\
\hline
Real time factor & 0.41 & 0.59 \\
\reviewchanges{Delayed event arrays} & $\unit[1.3]{\%}$ & $\unit[11.5]{\%}$\\
\bottomrule
\end{tabular}
\label{table:processing_time_stats}
\vspace{-0.4cm}
\end{table}
\section{Conclusion}
\label{sec:conclusions}
In this paper we introduce\reviewchanges{d} \finalsubmission{\sout{HoughCeption}Hough$^2$Map}, a high-speed real-time infrastructure mapping pipeline for utilization in a\reviewchanges{\sout{ future}} railway localization \reviewchanges{\sout{method}pipeline}. 
\reviewchanges{\sout{To achieve this, w}W}e \reviewchanges{developed\sout{presented}} a novel iterative \ac{HT} and \ac{NMS} for asynchronous event-based data as a method to detect typical railway infrastructure represented by vertical lines even at close range and high speeds.
\finalsubmission{\sout{HoughCeption}Hough$^2$Map} subsequently tracks detections over time in order to triangulate and map each \reviewchanges{\sout{one}object}. 

We validate our pipeline on two different datasets including real-life scenarios as well as a more challenging trajectory including occlusions, varying speeds and excess noise from close-by vegetation. 
Our experiments show that \finalsubmission{\sout{HoughCeption}Hough$^2$Map} is capable of reliably and accurately detecting and mapping poles and other infrastructure elements at close range even at high speeds, but\reviewchanges{, as expected, have difficulties with objects that\sout{struggles once these}} are \reviewchanges{\sout{placed}positioned} further away. 

%
%

We plan to expand this work into a full railway localization pipeline exploiting \reviewchanges{\sout{available ground truth}previously recorded} maps of \reviewchanges{\sout{both the tracks and}the} poles to perform accurate high-speed localization. 


\begin{acronym}
    \acro{GNSS}{Global Navigation Satellite System}
    \acro{IMU}{Inertial Measurement Unit}
    \acro{DVS}{Dynamic Vision Sensor}
    \acro{SLAM}{Simultaneous Localization and Mapping}
    \acro{DAVIS}{Dynamic and Active Vision Sensor}
    \acro{SNN}{Spiking Neural Network}
    \acro{DoF}{Degrees of Freedom}
    \acro{NMS}{Non-Maxima Suppression}
    \acro{ETCS}{European Train Control System}
    \acro{MAV}{Micro Aerial Vehicle}
    \acro{LSD}{Line Segment Detector}
    \acro{HT}{Hough transform}
    \acro{HS}{Hough space}
    \acro{FoV}{Field of View}
    \acro{RTK}{Real-Time Kinematic}
    \acro{RMSE}{Root Mean Square Error}
    \acro{INS}{Inertial Navigation System}
    \acro{EKF}{Extended Kalman Filter}
    \acro{DLT}{Direct Linear Transform}
    \acro{SVD}{Singular Value Decomposition}
\end{acronym}

\bibliographystyle{IEEEtran}
\bibliography{library,references}
\end{document}